\documentclass{article}

\usepackage{microtype}
\usepackage{graphicx}
\usepackage{subcaption}
\usepackage{booktabs} % for professional tables

\usepackage{hyperref}

\usepackage[preprint]{icml2026}

\usepackage{amsmath}
\usepackage{amssymb}
\usepackage{mathtools}
\usepackage{amsthm}

\usepackage{multirow}
\usepackage{tabularx}

\newtheorem{remark}{Remark}

\usepackage[capitalize,noabbrev]{cleveref}

\theoremstyle{plain}

\theoremstyle{definition}

\theoremstyle{remark}

\usepackage[textsize=tiny]{todonotes}

\icmltitlerunning{Function-Space Empirical Bayes Regularisation  with Large Vision–Language Model Priors}

\begin{document}

\twocolumn[
  \icmltitle{Function-Space Empirical Bayes Regularisation  \\
    with Large Vision–Language Model Priors}



  \icmlsetsymbol{equal}{*}

  \begin{icmlauthorlist}
    \icmlauthor{Pengcheng Hao}{yyy}
    \icmlauthor{Huaze Tang}{yyy}
    \icmlauthor{Ercan Engin Kuruoglu}{yyy}
    \icmlauthor{Wenbo Ding}{yyy}
  \end{icmlauthorlist}

  \icmlaffiliation{yyy}{Institute of Data and Information, Tsinghua Shenzhen International Graduate School, Shenzhen, China}
  %\icmlaffiliation{comp}{Company Name, Location, Country}

  \icmlcorrespondingauthor{Ercan Engin Kuruoglu}{kuruoglu@sz.tsinghua.edu.cn}
  %\icmlcorrespondingauthor{Firstname2 Lastname2}{first2.last2@www.uk}

  \icmlkeywords{Machine Learning, ICML}

  \vskip 0.3in
]



\printAffiliationsAndNotice{}  % no special notice (required even if empty)

\begin{abstract}
Bayesian deep learning (BDL) provides a principled framework for reliable uncertainty quantification by combining deep neural networks with Bayesian inference. A central challenge in BDL lies in the design of informative prior distributions that scale effectively to high-dimensional data. Recent functional variational inference (VI) approaches address this issue by imposing priors directly in function space; however, most existing methods rely on Gaussian process (GP) priors, whose expressiveness and generalisation capabilities become limited in high-dimensional regimes. In this work, we propose VLM-FS-EB, a novel function-space empirical Bayes regularisation framework, leveraging large vision-language models (VLMs) to generates semantically meaningful context points. These synthetic samples are then used VLMs for embeddings to construct expressive functional priors. Furthermore, the proposed method is evaluated against various baselines, and experimental results demonstrate that our method consistently improves predictive performance and yields more reliable uncertainty estimates, particularly in out-of-distribution (OOD) detection tasks and data-scarce regimes.
\end{abstract}

\section{INTRODUCTION}
% Deep neural network cannot qualify uncertainty % weight-space BNN training lacks informative prior and precise posterior
Deep neural networks (DNNs) have shown remarkable success across a wide range of tasks due to their high predictive accuracy. However, they lack the ability to quantify uncertainty, which is crucial in security-critical applications, such as autonomous driving, medical diagnosis, and financial decision-making. To address this limitation, one presents Bayesian neural networks (BNNs), which place prior distributions over model parameters and provide well-calibrated uncertainty estimation by approximating the posterior. Various Bayesian methods have been employed to perform the weight-space posterior inference, such as Markov Chain Monte Carlo (MCMC), Laplace approximation and variational inference (VI). However, there are two drawbacks of the weight-space training methods. 1) For computational tractability, the posterior distribution over weights is often approximated either by a simple Gaussian or a small set of discrete samples, which provides only a coarse representation of the true posterior. 2) Due to the complex relation between BNN predictions and their underlying weights, designing informative priors remains a significant challenge. 

% function-space regularisation and its drawbacks
To achieve interpretable prior,~\cite{sun2018functional} has proposed a function-space variational inference (FSVI) training approach, which defines prior prediction distributions and maximises a functional evidence lower bound (ELBO). However, there is no closed-form solution to the KL divergence over the infinite-dimensional prediction distributions. To resolve this problem, ~\cite{sun2018functional} approximates the supremum in the KL divergence by the spectral stein gradient estimator (SSGE)~\cite{shi2018spectral}, which incurs substantial computational cost and limits practical applicability. By contrast, the linearisation-based FSVI methods give a Gaussian prior assumption to the model parameters and approximate the functional distribution as a Gaussian process (GP). This yields a closed-form analytical solution for the KL divergence and enables efficient inference~\cite{immer2021improving, rudner2022tractable}. Although the KL divergence is well-defined since prior and variational posterior share the same support, the linear approximation can still cause training instability. Instead,~\cite{cinquin2024regularized} employs a pretrained GP prior and leverages the regularized KL divergence to guarantee a well well-defined objective. Nevertheless, linearizing the function mapping introduces large approximation errors and incurs substantial computational cost due to the required Jacobian matrix. In contrast, the function-space empirical Bayes (FS-EB) approach~\cite{rudner2023functionspace, rudner2024finetuning} avoids linear approximation and incorporates regularization in both parameter and function spaces. Nevertheless, a robust feature extractor for the task-relevant regions relies on pretraining data, and the functional regularization term necessitates context samples. These conditions are rarely satisfied in data-constrained settings, such as medical applications~\cite{kumari2025continual, guan2024federated}, limiting the applicability of this approach.

% For the limitation of Gaussian assumption, the heavy-tailed BNNs have been considered.
To overcome these limitations, we explore a VLM-FS-EB framework, leveraging large-scale vision-language models (VLMs) within the FS-EB framework. Such VLMs offer two key advantages in data-constrained settings: 1) they can generate diverse synthetic data; alleviating the need for extensive context samples. 2) they provide feature extractors with strong generalization, producing robust representations. The main contributions of this work consist in:
\begin{enumerate}
\item The VLM-FS-EB is introduced as a novel empirical Bayes framework that leverages large VLMs to synthesize informative context points in a controllable and data-free manner, thereby eliminating the reliance on external context data.
\item We replace task-specific feature extractors with a frozen, large embedding model to construct an expressive functional prior, bypassing costly domain-specific pretraining while inheriting rich semantic representations from foundation models.
\item  Comprehensive experiments across four real-world image benchmarks demonstrate the effectiveness of VLM-FS-EB under both standard and extreme data-scarce regimes, with consistent improvements over various baselines spanning function-space and parameter-space regularisation methods.
\end{enumerate}
The remainder of this paper is structured as follows: 
Section~\ref{sec: RW} reviews the related work. By contrast, Section~\ref{sec: pre} introduces the theoretical background, including LVMs for data generation and embeddings, and the FS-EB framework. Then, Section~\ref{sec: proposed} presents our proposed VLM-FS-EB method. Experimental evaluation is provided in Section~\ref{sec: evaluation} and Section~\ref{sec: conclusion} concludes the paper.

\section{Related Work}\label{sec: RW}
Our approach draws upon two key developments in modern machine learning: function-space Bayesian regularization and VLMs. In Section~\ref{sec: FSVI_R}, we review recent advances in FSVI methods which underpin our probabilistic formulation. By contrast, In Section~\ref{sec: LVM_ED} explaines how VLMs serve as powerful tools for both synthetic data generation and the construction of embedding-based priors.

\subsection{Function-space Variational Inference Regularisation}\label{sec: FSVI_R}
Early work on the FSVI in BNNs aimed to directly approximate posterior distributions over functions induced by neural networks.~\cite{wang2018function} introduces a flexible framework for approximate inference in Bayesian regression models, leveraging functional particle optimization–based VI. Also,~\citet{sun2018functional} shows that the KL divergence between stochastic processes can be expressed as the supremum of marginal KL divergences over all finite input sets, which enables approximation of the functional ELBO using finite measurement sets together with the SSGE method. However,~\cite{burt2021understanding} has shown that the function-space variational objective is generally ill-defined when combining neural network variational distributions with GP priors. Moreover, due to its high computational cost, the SSGE method is not suitable for high-dimensional data. To tackle intractable function-space objectives,~\cite{ma2019variational} introduces functional implicit stochastic process priors and an efficient VI framework to approximate the posterior. Also,~\cite{ma2021functional} proposes a FSVI method using stochastic process generators, where a grid-functional KL divergence enables a well-defined variational objective. By contrast, \cite{khan2019approximate} models a deep neural network (DNN) as a GP using Laplace and generalized Gauss–Newton (GGN) approximations, and derives a corresponding VI–based objective. Building on this framework, \cite{immer2021improving} extends the approach to non-Gaussian likelihoods, which however does not optimise the variance parameters. In comparison,~\cite{rudner2022tractable} introduces a fully BNN-based FSVI method, where the KL divergence is finite by defining the prior as the pushforward of a Gaussian prior distribution in weight space. Moreover, aiming for an interpretable prior, \cite{cinquin2024regularized} uses a pretrained GP prior and introduces a regularized KL divergence to guarantee a well-defined variational objective. However, these linearisation-based method introduces large computational costs and approximation errors. By contrast, the FS-EB approach~\cite{rudner2023functionspace} does not require linearisation  and leverages both parameter- and function-space priors. This work has been employed for tranfer learning~\cite{rudner2024fine} and subpopulation shift problems~\cite{rudner2024mind}. 

\subsection{Large Vision-Language Models for Embeddings and Data Generation}\label{sec: LVM_ED}

\paragraph{Synthetic Data Generation with Large Models for Uncertainty Quantification}~\citet{li2023synthetic} demonstrated that LLMs can generate synthetic training data, though effectiveness varies with task subjectivity. In the context of Bayesian deep learning, synthetic data generation capabilities of large models can be leveraged for data augmentation and uncertainty calibration. More recently, \citet{nadas2025synthetic} surveyed advances in LLM-based synthetic data generation, highlighting prompt-based generation and retrieval-augmented synthesis techniques. These approaches inspire our context sampling strategy, where VLMs generate semantically meaningful support points for functional priors. The ability to generate diverse, semantically coherent synthetic samples enables more robust uncertainty quantification by expanding the coverage of the input space during training and inference.

\paragraph{Vision-Language Models as Feature Extractors} Recent advances in large-scale VLMs such as CLIP~\cite{radford2021learning}, ALIGN~\cite{jia2021scaling}, and their successors have demonstrated remarkable capabilities in learning joint embeddings of visual and textual information. These models are pretrained on billions of image-text pairs, enabling them to capture rich semantic representations that generalize across diverse visual domains. The embedding spaces learned by VLMs exhibit strong alignment between visual and linguistic concepts, making them particularly suitable as informative priors for downstream tasks requiring robust uncertainty quantification. For example, \citet{dar2023analyzing} showed that transformer parameters can be interpreted directly in embedding space, revealing that pretrained representations encode structured semantic knowledge. The high-dimensional embedding spaces ($d_{\text{embed}} \sim 512\text{-}1024$) learned by VLMs provide rich geometric structures that naturally encode semantic similarities and hierarchical relationships. Building on this foundation, recent advances have explored the geometry of VLM embedding spaces for uncertainty-aware applications, showing that the semantic structure encoded in these spaces can be effectively leveraged for robust inference in safety-critical domains~\cite{silva2025full}. Recent work has focused on enhancing uncertainty estimation in VLM-based systems: \citet{morales2024bayesadapter} introduced BayesAdapter, which leverages Bayesian inference to estimate full probability distributions over adapter parameters instead of single point estimates, significantly improving calibration and selective classification performance in CLIP few-shot adaptation. Furthermore, \citet{zhou2025bayesian} proposed Bayesian test-time adaptation for VLMs, demonstrating that adapting both likelihood and prior distributions leads to more reliable uncertainty estimates in out-of-distribution (OOD) scenarios. Additionally, \citet{dar2023analyzing} showed that transformer parameters can be interpreted directly in embedding space, revealing that pretrained representations encode structured semantic knowledge. This finding is crucial for our work, as it suggests that VLM embeddings can serve as expressive functional priors that capture meaningful data structure beyond what traditional GP priors can achieve.

%\paragraph{Embedding Space Geometry and Transferability} \citet{dar2023analyzing} showed that transformer parameters can be interpreted directly in embedding space, revealing that pretrained representations encode structured semantic knowledge. This finding is crucial for our work, as it suggests that VLM embeddings can serve as expressive functional priors that capture meaningful data structure beyond what traditional GP priors can achieve. The high-dimensional embedding spaces ($d_{\text{embed}} \sim 512\text{-}1024$) learned by VLMs provide rich geometric structures that naturally encode semantic similarities and hierarchical relationships. Building on this foundation, recent advances have explored the geometry of VLM embedding spaces for uncertainty-aware applications, showing that the semantic structure encoded in these spaces can be effectively leveraged for robust inference in safety-critical domains~\cite{silva2025full}.

\section{Preliminary}\label{sec: pre}
In this section, we provide the necessary foundational concepts that underpin our proposed VLM-FS-EB framework. Specifically, Section~\ref{sec: VLM_SM} discusses the role of vision-language models (VLMs) in generating synthetic data and providing rich embeddings, and Section~\ref{sec: VLM_FSEB} explains the FS-EB framework, forming the basis of our method .

\subsection{Vision-Language Models for  Synthesis and Embedding}\label{sec: VLM_SM}
% \textcolor{blue}{Give some preliminary formulation description about Large model for embedding, kernel and data generation.}

In our framework, VLMs serve two distinct roles: (1) generating diverse synthetic images to expand input space coverage without real data. (2) providing frozen, semantic-rich embeddings for constructing expressive functional priors.  These two capabilities are respectively best supported by two dominant \textit{generative VLMs}~\cite{bordes2024introduction} and \textit{contrastive VLMs}.

\textbf{Generative VLMs.}
Modern generative VLMs adopt an \textit{encoder-decoder} architecture that connects a vision encoder to a LLM decoder~\cite{bordes2024introduction,wang2024qwen2vl}. Given an input image $x \in \mathbb{R}^{H \times W \times 3}$, a vision encoder (typically a Vision Transformer, ViT) first extracts visual features:
\begin{equation*}
    \mathbf{V} = \text{ViT}(x) = \{\mathbf{v}_1, \mathbf{v}_2, \ldots, \mathbf{v}_N\} \in \mathbb{R}^{N \times d_v},
\end{equation*}
where $N$ is the number of visual tokens and $d_v$ is the visual feature dimension. These visual tokens are then projected into the LLM's embedding space via a learnable adapter:
\begin{equation*}
    \mathbf{H}_{\text{vis}} = \mathcal{P}(\mathbf{V}) \in \mathbb{R}^{N' \times d_{\text{LLM}}},
\end{equation*}
where $\mathcal{P}$ is the projection layer, $N' \leq N$ is the number of projected visual tokens (often compressed for efficiency), and $d_{\text{LLM}}$ matches the LLM's hidden dimension. For a text query or instruction $q$, the text tokens are embedded as $\mathbf{H}_{\text{text}} = \text{Embed}(q) \in \mathbb{R}^{L \times d_{\text{LLM}}}$, where $L$ is the sequence length. The visual and textual representations are concatenated and fed into an autoregressive LLM decoder:
\begin{equation*}
    \mathbf{H} = [\mathbf{H}_{\text{vis}}; \mathbf{H}_{\text{text}}] \in \mathbb{R}^{(N' + L) \times d_{\text{LLM}}},
\end{equation*}
which generates responses token-by-token via next-token prediction:
\begin{equation*}
    p(y_t | y_{<t}, \mathbf{H}) = \text{Softmax}(\text{LLM}(\mathbf{H})_t \mathbf{W}_{\text{out}}),
\end{equation*}
where $\mathbf{W}_{\text{out}} \in \mathbb{R}^{d_{\text{LLM}} \times V}$ is the output projection matrix and $V$ is the vocabulary size. This multimodal generation capability is crucial for context point sampling in our proposed method, as it enables the synthesis of semantically meaningful data.

%Advanced generative VLMs introduce architectural innovations for enhanced visual understanding. \citet{wang2024qwen2vl} propose the \textit{Naive Dynamic Resolution} mechanism in Qwen2-VL, which dynamically processes images of varying resolutions into different numbers of visual tokens, and \textit{Multimodal Rotary Position Embedding (M-RoPE)} to effectively fuse positional information across modalities. This design allows flexible handling of high-resolution inputs and enables unified processing of both images and videos. %Unlike contrastive VLMs that produce fixed embeddings, generative VLMs support open-ended visual reasoning, detailed image captioning, visual question answering, and multimodal dialogue through their decoder-only language modeling framework.

\textbf{Contrastive VLMs.} 
For contrastive VLMS, such as CLIP \cite{radford2021learning}, ALIGN~\cite{jia2021scaling}, and their successors, its framework is typically consisted of an image encoder $\mathcal{E}_I$ and a text encoder $\mathcal{E}_T$. For an input image $x\in\mathbb{R}^{H\times W\times 3}$, where $H$ and $W$ are the height and width of the input image, the image encoder maps it to an embedding vector as 
\[\mathbf{z}_x = \mathcal{E}_I(x) \in \mathbb{R}^{d_{\text{embed}}}.\] 
Similarly, for a text description $t$, the text encoder produces 
\[\mathbf{z}_t = \mathcal{E}_T(t) \in \mathbb{R}^{d_{\text{embed}}},\] 
where both embeddings lie in a shared semantic space with dimension $d_{\text{embed}}$. The model is trained with a contrastive objective to maximize similarity between matched image-text pairs:
\begin{equation*}
    \mathcal{L}_{\text{contrastive}} = -\log \frac{\exp\left( \mathbf{z}_x^\top \mathbf{z}_t / \tau\right)}{\sum_{t'} \exp\left( \mathbf{z}_x^\top \mathbf{z}_{t'} / \tau\right)},
\end{equation*}
where $\tau$ is a temperature parameter and the summation runs over negative text samples. This dual-encoder architecture~\cite{radford2021learning,jia2021scaling} can learn semantic alignment and produces fixed-dimensional embeddings, enabling efficient similarity computation—a property we leverage to measure coherence and construct our VLM-FS-EB prior.

\subsection{Function-Space Empirical Bayes Regularisation}\label{sec: VLM_FSEB}
This section explains the FS-EB framework proposed in~\cite{rudner2023functionspace,rudner2024finetuning}, which combines parameter-space and function-space regularisation methods and constitutes the foundation of our proposed method. 

Consider a supervised learning task with a dataset consisting of $N$ independent and identically distributed samples, denoted by \[\mathcal{D} = \left\{ x^{(n)}_{\mathcal{D}}, y^{(n)}_{\mathcal{D}} \right\}_{n=1}^{N} = \left( \mathbf{x}_{\mathcal{D}}, \mathbf{y}_{\mathcal{D}} \right).\] 
Here, each input $x^{(n)}_{\mathcal{D}}$ belongs to the input space $\mathcal{X} \subseteq \mathbb{R}^{D}$, and its corresponding target $y^{(n)}_{\mathcal{D}}$ lies in the target space $\mathcal{Y}$. 
For regression problems, the target space $\mathcal{Y} \subseteq \mathbb{R}^{K}$, while 
for classification tasks involving $K$ classes, the target space is a subset of the set of $K$-dimensional binary vectors, i.e., $\mathcal{Y} \subseteq \{0, 1\}^{K}$.

The FS-EB begins by defining an auxiliary inference problem over a set of context points $\mathbf{x}_c = \{x_c^{(1)}, \ldots, x_c^{(M)}\}$ with corresponding labels $\mathbf{y}_c$. Given a neural network $f(\cdot;\theta)$ and a prior over parameters $p(\theta)$, the auxiliary posterior is defined as
\begin{equation}
    p(\theta \mid \mathbf{y}_c, \mathbf{x}_c) \propto p(\mathbf{y}_c \mid \mathbf{x}_c, \theta;f)\, p(\theta).
\end{equation}
This posterior serves as an empirical prior for the main inference task. To define the likelihood $p(\mathbf{y}_c \mid \mathbf{x}_c, \theta;f)$, FS-EB employs a stochastic linear model over a feature extractor $h(\cdot;\phi_0)$,
\[
z(x) \overset{\cdot}{=} h(x;\phi_0)\Psi + \epsilon,
\]
where $\phi_0$ is the parameter vector, $\Psi \sim \mathcal{N}(\mathbf{0}, \tau_1 I)$ and $\epsilon \sim \mathcal{N}(\mathbf{0}, \tau_2 I)$. Also, $ I $  is the identity matrix and  $ \tau_1, \tau_2 \in \mathbb{R}^{+} $  are variance hyperparameters. This model induces a Gaussian distribution over function evaluations at the context points,
\begin{equation*}
    p(z \mid \mathbf{x}_c) = \mathcal{N}\big(z; \mathbf{0},  K(\mathbf{x}_c, \mathbf{x}_c)\big),
\end{equation*}
with covariance matrix
\[
K(\mathbf{x}_c, \mathbf{x}_c) = \tau_1 h(\mathbf{x}_c;\phi_0) h(\mathbf{x}_c;\phi_0)^\top + \tau_2I.
\]

Viewing this distribution as a likelihood over neural network outputs, FS-EB defines
\begin{equation}\label{eq: FSEB likelihood}
  p_k(\mathbf{y}_c \mid \mathbf{x}_c, \theta;f) =
\mathcal{N}\big(\mathbf{y}_c; [f(\mathbf{x}_c;\theta)]_k,\,
 K(\mathbf{x}_c, \mathbf{x}_c)\big), 
\end{equation}

where $\mathbf{y}_c = \mathbf{0}$ and $[f(\mathbf{x}_c;\theta)]_k$ is the $k$-th component of $f(\mathbf{x}_c;\theta)$. Then we have
\begin{equation}\label{eq:FSEB prior}
  p(\theta \mid \mathbf{y}_c, \mathbf{x}_c) \propto p(\theta)
\prod_{k=1}^{K}p_k(\mathbf{y}_c \mid \mathbf{x}_c, \theta;f)  
\end{equation}

Using this auxiliary posterior as an empirical prior, the posterior distribution $p(\theta \mid \mathbf{y}_{\mathcal{D}}, \mathbf{x}_{\mathcal{D}})$ is defined by
\begin{equation}\label{eq: FSEB posterior}
p(\theta \mid \mathbf{y}_{\mathcal{D}}, \mathbf{x}_{\mathcal{D}})
\propto p(\mathbf{y}_{\mathcal{D}} \mid \mathbf{x}_{\mathcal{D}}, \theta)\,
p(\theta \mid \mathbf{y}_c, \mathbf{x}_c).
\end{equation}
To approximate the posterior distribution, \cite{rudner2023functionspace} introduces estimation methods based on maximum a posteriori (MAP) and VI.

\begin{figure*}[t]
  \begin{center}
    \includegraphics[width=1.0\textwidth]{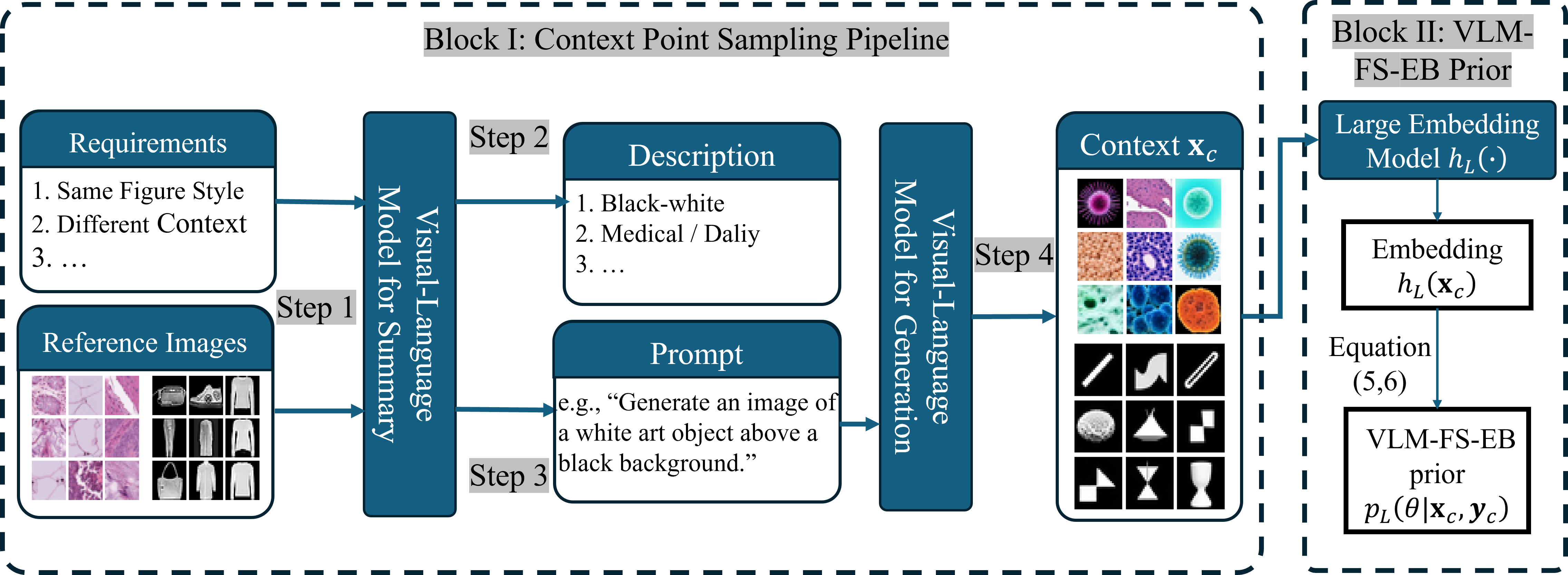}
    \caption{
      Illustration of the VLM-FS-EB prior framework. Block I shows the pipeline for generating context points $\mathbf{x}_c$ via VLMs. Block II depicts how these synthetic context points are used to construct the VLM-FS-EB functional prior via a VLM for embeddings $h_{L}(\cdot)$.
    }
    \label{fig:vlm-fs-eb}
  \end{center}
  \vspace{-10pt}
\end{figure*}

\section{Proposed Method: VLM-FS-EB}\label{sec: proposed}
In the FS-EB framework described above, the function-space prior is induced by a stochastic linear model defined over a fixed feature extractor $h(\cdot;\phi_0)$. Although this formulation is general, both the choice of context points $\mathbf{x}_c$ and the quality of $h(\cdot;\phi_0)$ critically influence the expressiveness and generalization capability of the resulting prior. In data-scarce settings, suitable context points are often difficult to obtain, and obtaining a high-quality $h(\cdot;\phi_0)$ is equally challenging, as it typically relies on large-scale, high-quality pretraining data that may not be available.

This section explains the VLM-FS-EB, a framework that eliminates the need for additional data and demonstrates strong generalization even under extreme data scarcity. Section~\ref{sec: context sampling} introduces our method for sampling context points using a large VLM. Section~\ref{sec: VLM-FS-EB prior} constructs an expressive functional prior based on a VLM for embeddings. Also, the posterior approximation via Monte Carlo (MC) dropout is detailed in Section~\ref{sec: Posterior approximation}.

\subsection{Context Point Sampling via Large Vision-Language Models}\label{sec: context sampling}

% The key challenge is: \textit{where do these context points come from?} Traditional approaches either sample them from additional datasets or randomly select them from training batches, both of which have limitations—external data may not be available, and random sampling may not adequately cover the semantically meaningful variations in the input space. 

Our solution leverages the generative capabilities of large VLMs to synthesize diverse, task-relevant context points in a controllable manner. Our context point sampling strategy is illustrated in Figure~\ref{fig:vlm-fs-eb}: Block I. The generation pipeline operates as follows:
\textbf{Step 1: Input Specification.} The pipeline takes two types of inputs: (1) \textit{Requirements}: textual constraints specifying desired properties such as ``maintain the same visual style'' or ``generate variations with the same style''; (2) \textit{Reference Images}: a small set of representative samples from the training data that serve as visual anchors for the generation process.
\textbf{Step 2: Multi-modal Understanding.} These inputs are processed by a VLM (e.g., Qwen-VL), which acts as an intelligent summarizer. The VLM simultaneously analyzes the visual content of the reference images and interprets the textual requirements, producing a unified understanding of both modalities. For instance, given reference images of handwritten digits and the requirement ``generate diverse digit styles,'' the VLM identify key description including image style and usage.
\textbf{Step 3: Prompt Construction.} Based on this multi-modal understanding, the VLM automatically constructs a detailed text prompt that captures both the semantic content and stylistic attributes of the desired outputs.
\textbf{Step 4: Image Synthesis.} The constructed prompt is then fed into a generative VLM to synthesize new images. By varying the prompts systematically (e.g., changing the digit class, stroke style, or background), the proposed framework can generate a diverse set of context points that comprehensively cover the semantically meaningful variations in the input space.
\vspace{-10pt}
\begin{remark}
    This approach requires no external data beyond the given few-shot training samples, yet enables semantically meaningful exploration of the context space through controllable generation. By ``semantically meaningful,'' we mean that the generated samples capture high-level concepts and variations (e.g., object categories, visual styles, poses) rather than arbitrary pixel-level noise. The synthesized samples serve directly as context points to inform the empirical Bayes prior without requiring manual inspection, filtering, or annotation. These properties enable our method to generalize well in real-world settings where data is scarce or manual annotation is expensive.
\end{remark}

\subsection{Functional Priors with Large Vision-Language Embedding Models}\label{sec: VLM-FS-EB prior}
Following the sampling processing of context points, we design the VLM-FS-EB prior as show in Figure~\ref{fig:vlm-fs-eb}:Block II. Specifically, we replace the task-specific feature extractor
$h(\cdot;\phi_0)$ with a large embedding model and obtain the embedding vector $h_{L}(\mathbf{x}_c)$. According to equation~(\ref{eq: FSEB likelihood}), the likelihood over $[f(\mathbf{x}_c;\theta)]_k$ given context inputs $\mathbf{x}_c $ can be modeled as a multivariate Gaussian distribution:
\begin{equation}\label{eq: VLM-FSEB likelihood}
  p_L^k(\mathbf{y}_c \mid \mathbf{x}_c, \theta;f) =
\mathcal{N}\big(\mathbf{y}_c; [f(\mathbf{x}_c;\theta)]_k,\,
 K_L(\mathbf{x}_c, \mathbf{x}_c)\big), 
\end{equation}
where $\mathbf{y}_c=\mathbf{0}$ and the covariance matrix
\[
K_{L}(\mathbf{x}_c, \mathbf{x}_c) = \tau_1 \, h_{L}(\mathbf{x}_c) h_{L}(\mathbf{x}_c)^\top + \tau_2 \, I
\]
is constructed from the frozen large-model embeddings   $  h_L(\mathbf{x}_c)$. Using this likelihood, the corresponding 
auxiliary posterior $p_L(\theta \mid \mathbf{y}_c, \mathbf{x}_c)$ can be constructed in the same manner as euqation (\ref{eq:FSEB prior}), i.e.,
\begin{equation}\label{eq:VLM-FSEB prior}
  p_L(\theta \mid \mathbf{y}_c, \mathbf{x}_c) \propto p(\theta)
\prod_{k=1}^{K}p_L^k(\mathbf{y}_c \mid \mathbf{x}_c, \theta;f)  
\end{equation}
and serves as an empirical prior for the main task. 
\begin{remark}
  Leveraging the general-purpose representations of a large frozen  embedding model $h_{\text{L}}(\cdot)$, we bypass task-specific feature pretraining and hence mitigate the requirement for extensive domain-specific data. 
\end{remark}

\subsection{Scalable Posterior Inference via MC Dropout}\label{sec: Posterior approximation}

We now describe how posterior inference is performed under the proposed VLM-FS-EB prior. According to equation~(\ref{eq: FSEB posterior}), the posterior distribution can be written as
\[
p_L(\theta \mid \mathbf{y}_{\mathcal{D}}, \mathbf{x}_{\mathcal{D}})
\propto p(\mathbf{y}_{\mathcal{D}} \mid \mathbf{x}_{\mathcal{D}}, \theta)\,
p_L(\theta \mid \mathbf{y}_c, \mathbf{x}_c).
\]
As the posterior $p_L(\theta \mid \mathbf{x}_{\mathcal{D}},
\mathbf{y}_{\mathcal{D}})$ is intractable, we adopt VI by introducing a variational distribution  $ q(\theta) $ and seek to
minimize the KL divergence
\[
\min_{q \in \mathcal{Q}} D_{\mathrm{KL}}
\big(q(\theta) \,\|\, p_L(\theta \mid \mathbf{x}_{\mathcal{D}},
\mathbf{y}_{\mathcal{D}})\big),
\]
where $\mathcal{Q}$ denotes a tractable variational family. This optimization problem is equivalent to maximizing ELBO,
\begin{align}\label{eq:ELBO}
  \mathcal{L}(\theta)
& = \mathbb{E}_{q(\theta)}
\big[p(\mathbf{y}_{\mathcal{D}} \mid \mathbf{x}_{\mathcal{D}}, \theta)\big] \nonumber\\
& \quad - D_{\mathrm{KL}}\big(q(\theta) \,\|\, p_L(\theta \mid \mathbf{y}_c, \mathbf{x}_c)\big)  
\end{align}
The first term is the expected data log-likelihood, and the second regularises the variational posterior toward the VLM-FS-EB empirical prior.

\paragraph{Variational approximation via MC dropout.}
To obtain a tractable approximation to the variational posterior $q(\theta)$, we employ MC dropout~\cite{gal2016dropout}. This scheme  implicitly defines  $ q(\theta) $  over the network weights, and each stochastic forward pass via a random dropout mask corresponds to sampling a realization  $ \theta^{(s)} \sim q(\theta) $ , where  $ \theta^{(s)} $  denotes the effective parameters under the  $s$-th dropout mask. %Under the standard choice of a zero-mean Gaussian prior, the KL divergence term in~(\ref{eq:ELBO}) reduces to an  $\ell_2$  weight decay. 
The resulting stochastic objective can be approximated as
\[
\begin{aligned}
\hat{\mathcal{L}}(\theta)
=&\;
\frac{1}{S} \sum_{s=1}^{S}
\Bigg[
\log p(\mathbf{y}_{\mathcal{D}} \mid \mathbf{x}_{\mathcal{D}}, \theta^{(s)})
\\
&\qquad
+
\sum_{k=1}^{K}
\log p_L^k(\mathbf{y}_c \mid \mathbf{x}_c, \theta^{(s)}; f)
\Bigg]
- \lambda \lVert \theta \rVert_2^2,
\end{aligned}
\]
where $S$ denotes the number of MC samples and $\lambda$ controls the strength of regularisation. At inference time, predictions are obtained by averaging the outputs of multiple stochastic forward passes (with dropout active), which corresponds to MC integration under the variational posterior  $q(\theta)$ .

\begin{table*}[!tbhp]
\centering
\caption{In-distribution prediction performance under full training data. 
\textbf{Best}  and \underline{second best} results are highlighted}
\label{tab:main_result_1}
\setlength{\tabcolsep}{9pt} 
\resizebox{1.0\textwidth}{!}{
\begin{tabular}{l l c c c c c c}
\toprule
Metric & Dataset  & VLM-FS-EB & FS-EB & GFSVI & Dropout & MFVI & MAP \\
\midrule

\multirow{3}{*}{ACC $\uparrow$}
& MNIST &99.24 $ \pm $ 0.105 & 99.11  $\pm$ 0.048 & \underline{99.30  $\pm$ 0.036} & \textbf{99.32  $\pm$ 0.035} & 99.15  $\pm$ 0.095 & 99.09  $\pm$ 0.062 \\
& FMNIST&91.96  $ \pm $  0.187 & 90.46  $\pm$ 0.171 & \underline{92.08  $\pm$ 0.096} & \textbf{92.29  $\pm$ 0.189} & 90.28  $\pm$ 0.281 & 91.87  $\pm$ 0.242 \\
& CIFAR-10& \textbf{83.73  $ \pm $  0.549} & 75.67  $\pm$ 0.333 & 72.88  $\pm$ 0.742 & \underline{81.57  $\pm$ 0.515} & 76.02  $\pm$ 0.447 & 72.98  $\pm$ 0.818  \\
& PathMNIST & 87.23  $  \pm  $  1.858  & \underline{87.69  $\pm$ 0.945} & 84.81  $\pm$ 0.772 & 86.54  $\pm$ 1.324 & 86.96  $\pm$ 1.675 & \textbf{87.72  $\pm$ 2.446}    \\
\midrule

\multirow{3}{*}{ECE $\downarrow$}
& MNIST& 0.014  $ \pm $  0.004 & 0.019  $\pm$ 0.001 & \textbf{0.003  $\pm$ 0.000} & 0.012  $\pm$ 0.001 & \underline{0.008  $\pm$ 0.001} & \textbf{0.003  $\pm$ 0.001} \\
& FMNIST& 0.032  $ \pm $  0.002 & 0.023  $\pm$ 0.003 & \textbf{0.008  $\pm$ 0.002} & 0.038  $\pm$ 0.004 & \underline{0.011  $\pm$ 0.003} & 0.012  $\pm$ 0.003 \\
& CIFAR-10 & 0.090  $ \pm $  0.006& 0.102  $\pm$ 0.004 & \textbf{0.024  $\pm$ 0.012} & 0.088  $\pm$ 0.007 & 0.081  $\pm$ 0.005 & \underline{0.038  $\pm$ 0.012} \\

& PathMNIST  & \underline{0.038  $\pm$  0.009}  & \textbf{0.019 $\pm$ 0.006} & 0.039 $\pm$ 0.021 & 0.041 $\pm$ 0.007 & \textbf{0.019 $\pm$ 0.004} & 0.048 $\pm$ 0.013 \\
\midrule

\multirow{3}{*}{NLL $\downarrow$}
& MNIST & 0.033  $ \pm $  0.006 & 0.040  $\pm$ 0.001 & \textbf{0.022  $\pm$ 0.002} & 0.029  $\pm$ 0.002 & 0.029  $\pm$ 0.001 & \underline{0.026  $\pm$ 0.001} \\
& FMNIST & {0.232  $ \pm $  0.004} & 0.268  $\pm$ 0.005 & \textbf{0.227  $\pm$ 0.002} & \textbf{0.227  $\pm$ 0.004} & 0.269  $\pm$ 0.006 & \underline{0.231  $\pm$ 0.005} \\
& CIFAR-10 & \textbf{0.525  $ \pm $  0.013} & 0.761  $\pm$ 0.008 & 0.803  $\pm$ 0.018 & \underline{0.577  $\pm$ 0.017} & 0.724  $\pm$ 0.013 & 0.794  $\pm$ 0.019 \\
& PathMNIST  & 0.412  $  \pm  $  0.069  & \textbf{0.378 $\pm$ 0.030} & 0.464 $\pm$ 0.019 & 0.467 $\pm$ 0.043 & \underline{0.391 $\pm$ 0.035} & 0.435 $\pm$ 0.093 \\
\bottomrule
\end{tabular}
}
\end{table*}

\begin{table*}[!tbhp]
\centering
\caption{OOD detection performance under full training data. \textbf{Best}  and \underline{second best} results are highlighted}
\label{tab:main_result_2}
\setlength{\tabcolsep}{9pt} 
\resizebox{1.0\textwidth}{!}{
\begin{tabular}{l l c c c c c c}
\toprule
In & OOD Dataset & VLM-FS-EB &  FS-EB & GFSVI & Dropout & MFVI & MAP \\
\midrule

\multirow{4}{*}{MNIST}
& FMNIST & \textbf{99.99  $ \pm $  0.005} & \underline{99.04  $\pm$ 0.140} & 98.63  $\pm$ 0.591 & 98.59  $\pm$ 0.351 & 98.60  $\pm$ 0.231 & 98.56  $\pm$ 0.255 \\
& NotMNIST & \textbf{100.00  $ \pm $  0.000} & {95.43  $\pm$ 0.670} & 91.41  $\pm$ 3.851 & \underline{96.86  $\pm$ 0.286} & 94.06  $\pm$ 0.777 & 95.42  $\pm$ 0.988 \\
& MNIST-c & \textbf{92.12  $ \pm $  0.956} & \underline{85.26  $\pm$ 1.087} & 84.08  $\pm$ 3.327 & 84.60  $\pm$ 0.679 & 82.69  $\pm$ 0.943 & 82.30  $\pm$ 0.743 \\

\midrule

\multirow{2}{*}{FMNIST}
& MNIST & \underline{85.28  $ \pm $  1.831} & 83.81  $\pm$ 2.929 & \textbf{88.54  $\pm$ 1.841} & 81.22  $\pm$ 1.717 & {77.48  $\pm$ 2.726} & 78.40  $\pm$ 2.310 \\
& NotMNIST & \textbf{86.72  $ \pm $  1.588} & 76.71  $\pm$ 3.550 & 75.33  $\pm$ 1.890 & \underline{80.03  $\pm$ 1.611} & {70.57  $\pm$ 4.324} & 69.49  $\pm$ 1.552 \\
\midrule

\multirow{3}{*}{CIFAR-10}
& SVHN & \textbf{85.64  $ \pm $  1.409} & 81.90  $\pm$ 1.967 & 76.95  $\pm$ 0.817 & 81.34  $\pm$ 3.330 & \underline{82.16  $\pm$ 1.849} & 79.51  $\pm$ 1.806 \\
& CIFAR-10C0 & \textbf{73.97  $ \pm $  1.759} & \underline{67.96  $\pm$ 1.276} & 60.85  $\pm$ 1.008 & 67.29  $\pm$ 1.809 & 67.33  $\pm$ 1.419 & 63.50  $\pm$ 1.515 \\
& CIFAR-10C2 & \textbf{64.31  $ \pm $  1.220} & \underline{62.65  $\pm$ 0.926} & 56.52  $\pm$ 0.658 & 62.06  $\pm$ 0.782 & 62.44  $\pm$ 0.751 & 58.98  $\pm$ 0.934 \\
& CIFAR-10C4 & {54.45  $ \pm $  0.539} & \textbf{55.38  $\pm$ 0.432} & 52.49  $\pm$ 0.349 & \underline{54.59  $\pm$ 0.409} & 55.17  $\pm$ 0.420 & 53.41  $\pm$ 0.404 \\

\midrule
\multirow{1}{*}{PathMNIST}
& BloodMNIST & \textbf{87.63  $  \pm  $  10.509}   & 82.87 $\pm$ 10.357 & 74.70 $\pm$ 6.596 & 70.39 $\pm$ 14.816 & \underline{83.99 $\pm$ 8.922} & 73.09 $\pm$ 13.003  \\

\bottomrule
\end{tabular}
}
\end{table*}

\begin{table*}[!tbhp]
\centering
\caption{In-distribution prediction performance under partial training data settings.
\textbf{Best}  and \underline{second best} results are highlighted}
\label{tab:main_result_3}
\setlength{\tabcolsep}{9pt} 
\resizebox{1.0\textwidth}{!}{
\begin{tabular}{l l c c c c c c}
\toprule
Metric & Dataset  & VLM-FS-EB & FS-EB & GFSVI & Dropout & MFVI & MAP \\
\midrule

\multirow{3}{*}{ACC $\uparrow$}
& MNIST & \textbf{98.98  $ \pm $  0.086} & \underline{98.86  $\pm$ 0.081} & 98.71  $\pm$ 0.111 & 98.77  $\pm$ 0.128  & 98.05  $\pm$ 0.462 & 98.51  $\pm$ 0.071 \\
& FMNIST& \textbf{90.19  $ \pm $  0.374} & 89.09  $\pm$ 0.474 & 89.73  $\pm$ 0.234 & \underline{90.07  $\pm$ 0.268}  & 88.81  $\pm$ 0.999 & 89.18  $\pm$ 0.344 \\
& CIFAR-10  & \textbf{74.93  $ \pm $  0.525} & 58.34  $\pm$ 1.256 & 54.33  $\pm$ 2.530 & \underline{70.45  $\pm$ 5.755} & 52.95  $\pm$ 1.529 & 57.81  $\pm$ 1.816  \\
& PathMNIST &\underline{84.56  $  \pm  $  2.159} & 82.26  $ \pm $  1.328 & 78.15  $ \pm $  1.462 & \textbf{85.69  $ \pm $  1.085} & 82.27  $ \pm $  1.554 & 79.44  $ \pm $  2.979 \\
\midrule

\multirow{3}{*}{ECE $\downarrow$}
& MNIST& 0.017  $ \pm $  0.002 & 0.016  $\pm$ 0.001 & \textbf{0.003  $\pm$ 0.002} &  0.022  $\pm$ 0.003 & 0.011  $\pm$ 0.002 & \underline{0.004  $\pm$ 0.001} \\
& FMNIS& 0.022  $ \pm $  0.003 & \underline{0.013  $\pm$ 0.003} &0.029  $\pm$ 0.010  & 0.031  $\pm$ 0.002  & \textbf{0.008  $\pm$ 0.002} & 0.023  $\pm$ 0.006 \\
& CIFAR-10  & 0.114  $ \pm $  0.013 & 0.083  $\pm$ 0.010 & \underline{0.043  $\pm$ 0.018} & 0.075  $\pm$ 0.006 & \textbf{0.019  $\pm$ 0.015} & 0.069  $\pm$ 0.021  \\
& PathMNIST  & \underline{0.027  $  \pm  $  0.009}  & 0.048  $ \pm $  0.012 & \textbf{0.023  $\pm $  0.015} & \underline{0.027  $ \pm $  0.008} & 0.044  $ \pm $  0.012 & 0.104  $ \pm $  0.022 \\  
\midrule

\multirow{3}{*}{NLL $\downarrow$}
& MNIST & \underline{0.044  $ \pm $  0.004} & 0.045  $\pm$ 0.001 & \textbf{0.039  $\pm$ 0.003}  & 0.053  $\pm$ 0.005  & 0.066  $\pm$ 0.016 & 0.046  $\pm$ 0.002 \\
& FMNIST & \textbf{0.284  $ \pm $  0.010} & 0.306  $\pm$ 0.010 & 0.320  $\pm$ 0.021  & \underline{0.286  $\pm$ 0.003}  & 0.313  $\pm$ 0.019 & 0.310  $\pm$ 0.006 \\
& CIFAR-10  & \textbf{0.794  $ \pm $  0.017} & 1.247  $\pm$ 0.027 & 1.296  $\pm$ 0.058 & \underline{0.869  $\pm$ 0.145} & 1.326  $\pm$ 0.035 & 1.218  $\pm$ 0.037   \\
& PathMNIST & \textbf{0.489  $  \pm  $  0.074} & 0.585  $ \pm $  0.040 & 0.637  $ \pm $  0.050 &  \underline{0.535  $ \pm $  0.087} & 0.578  $ \pm $  0.031 & 0.884  $ \pm $  0.157    \\
\bottomrule
\end{tabular}
}
\end{table*}

\begin{table*}[!tbhp]
\centering
\caption{OOD detection performance under partial training data settings.
\textbf{Best}  and \underline{second best} results are highlighted}
\label{tab:main_result_4}
\setlength{\tabcolsep}{9pt} 
\resizebox{1.0\textwidth}{!}{
\begin{tabular}{l l c c c c c c}
\toprule
In & OOD Dataset & VLM-FS-EB &  FS-EB & GFSVI & Dropout & MFVI & MAP \\
\midrule

\multirow{4}{*}{MNIST}
& FMNIST  & \textbf{99.99  $ \pm $  0.006}  & \underline{98.91  $\pm$ 0.171}  & 97.49  $\pm$ 0.709 & 98.25  $\pm$ 0.326   & 96.34  $\pm$ 1.425 & 97.62  $\pm$ 0.514 \\
& NotMNIST & \textbf{100.00  $ \pm $  0.002} & 95.60  $\pm$ 0.662 & 89.22  $\pm$ 4.893 & \underline{96.13  $\pm$ 0.264}  & 90.23  $\pm$ 3.054  & 93.25  $\pm$ 1.346 \\
& MNIST-c & \textbf{91.55 $ \pm $ 0.918}  & \underline{84.21  $\pm$ 0.958} & 75.90  $\pm$ 2.523 & 82.00  $\pm$ 0.643  & 79.19  $\pm$ 1.683 & 80.36  $\pm$ 0.916 \\
\midrule

\multirow{2}{*}{FMNIST}
& MNIST & \textbf{86.18  $ \pm $  2.151} & 83.88  $\pm$ 4.176 & \underline{85.38  $\pm$ 2.635} & 83.12  $\pm$ 2.363  & 78.02  $\pm$ 6.009 & 71.82  $\pm$ 1.986 \\
& NotMNIST & \textbf{85.84  $ \pm $  1.338} & 77.45  $\pm$ 3.198 & 77.06  $\pm$ 1.796  & \underline{80.34  $\pm$ 2.700}  & 72.78  $\pm$ 3.409 & 66.20  $\pm$ 2.374 \\
\midrule

\multirow{3}{*}{CIFAR-10}
& SVHN & \textbf{79.88  $ \pm $  2.188} & 72.09  $\pm$ 1.314 & 71.40  $\pm$ 4.152 & 73.62  $\pm$ 3.311 & 70.59  $\pm$ 2.574 & \underline{74.61  $\pm$ 3.468}   \\
& CIFAR-10C0 & \textbf{71.71  $ \pm $  1.918} & 56.60  $\pm$ 0.992 & 56.28  $\pm$ 2.136 & \underline{61.89  $\pm$ 2.608} & 56.17  $\pm$ 0.913 & 57.63  $\pm$ 1.788  \\
& CIFAR-10C2 & \textbf{62.09  $ \pm $  1.434} & 54.12  $\pm$ 0.665 & 53.77  $\pm$ 1.365 & \underline{57.61  $\pm$ 1.884} & 53.99  $\pm$ 0.552 & 54.82  $\pm$ 1.082 \\
& CIFAR-10C4 & \textbf{53.28  $ \pm $  0.559} & 51.86  $\pm$ 0.376 & 51.48  $\pm$ 0.720 & \underline{52.88  $\pm$ 0.793} & 51.78  $\pm$ 0.344 & 52.05  $\pm$ 0.484 \\
\midrule
\multirow{1}{*}{PathMNIST}
& BloodMNIST & \textbf{99.66  $  \pm  $  0.137}  & 83.32  $ \pm $  10.213 & 61.59  $ \pm $  6.108 & 42.85  $ \pm $  25.149 & \underline{83.67  $ \pm $  8.773} & 58.26  $ \pm $  17.689 \\

\bottomrule
\end{tabular}
}
\end{table*}

% \begin{table*}[!hp]
\begin{table*}[!tbhp]
\centering
\caption{Ablation study on VLM-generated embeddings and context points. 
For MNIST, OOD1 = FashionMNIST and OOD2 = NotMNIST; for FashionMNIST, OOD1 = MNIST and OOD2 = NotMNIST; and for CIFAR-10, OOD1 = SVHN and OOD2 = CIFAR-10C0.
The AUROC OOD scores are based on  MSP and predictive entropy). 
\textbf{Best} and \underline{second best} results are highlighted.
}
\label{tab:main_result_5}
\setlength{\tabcolsep}{9pt} 
\resizebox{1.0\textwidth}{!}{
\begin{tabular}{l l c c c c c c}
\toprule
 & Variants  & ACC$\uparrow$ & NLL$\downarrow$ & $\mathrm{OOD1}_{max} \uparrow$ & $\mathrm{OOD2}_{max}\uparrow$ & $\mathrm{OOD1}_{entropy}\uparrow$ & $\mathrm{OOD2}_{entropy}\uparrow$ \\
\midrule

\multirow{2}{*}{MNIST }
& VLM-VLM & 99.24  $ \pm $  0.105 & \textbf{0.033  $ \pm $  0.006} & \textbf{99.99  $ \pm $  0.005} & \textbf{100.00  $ \pm $  0.000 }& \textbf{100.00  $ \pm $  0.003} & \textbf{100.00  $ \pm $  0.000}\\
& VLM-Rand &\textbf{99.35  $ \pm $  0.064}  & 0.034  $ \pm $  0.003&99.28  $ \pm $  0.183 & 98.44  $ \pm $  0.317 &99.49  $ \pm $  0.156 & 98.63  $ \pm $  0.315  \\
& Rand-VLM & 99.21  $ \pm $  0.094 & 0.035  $ \pm $  0.005& 99.98  $ \pm $  0.012 & 99.99  $ \pm $  0.002 & 99.99  $ \pm $  0.006 & \textbf{100.00  $ \pm $  0.001}  \\
\midrule

\multirow{2}{*}{FMNIST }
& VLM-VLM &91.96  $ \pm $  0.187 &\textbf{0.232  $ \pm $  0.004} &\textbf{85.28  $ \pm $  1.831} &\textbf{86.72  $ \pm $  1.588} &\textbf{88.89  $  \pm  $  1.884} & \textbf{89.99  $  \pm  $  1.513}\\
& VLM-Rand & \textbf{92.05  $ \pm $  0.334}  & \textbf{0.232  $ \pm $  0.007}& 82.61  $ \pm $  2.029 & 80.73  $ \pm $  2.979 &85.40  $ \pm $  2.129 & 83.49  $ \pm $  2.787 \\
& Rand-VLM & 91.80  $ \pm $  0.272  & 0.235  $ \pm $  0.008 &79.80  $ \pm $  2.488 & 80.07  $ \pm $  2.304 &82.21  $ \pm $  2.561 & 82.48  $ \pm $  2.358 \\
\midrule

\multirow{2}{*}{CIFAR-10 }
& VLM-VLM &83.73  $ \pm $  0.549 & \textbf{0.525  $ \pm $  0.013} & \textbf{85.64  $ \pm $  1.409} & \textbf{73.97  $ \pm $  1.759} & \textbf{86.32  $ \pm $  1.550} & \textbf{75.66  $ \pm $  1.776} \\
& VLM-Rand & \textbf{83.88  $ \pm $  0.325}  & 0.531  $ \pm $  0.007  & 80.92  $ \pm $  1.840 & 69.73  $ \pm $  1.274
& 80.38  $ \pm $  2.345  & 70.71  $ \pm $  1.357 \\
& Rand-VLM & 81.58  $ \pm $  0.777  & 0.579  $ \pm $  0.023 & 79.77  $ \pm $  3.712 & 68.29  $ \pm $  1.234 & 80.43  $ \pm $  4.059 & 69.72  $ \pm $  1.348 \\
\bottomrule
\end{tabular}
}
\end{table*}

\section{Empirical Evaluation} \label{sec: evaluation}

\textbf{Baselines.} 
We compare our method against two FSVI approaches: FS-EB~\cite{rudner2023functionspace} and generalized function-space variational inference (GFSVI)~\cite{cinquin2024regularized}. In addition, we evaluate two parameter-space VI methods, namely MC Dropout~\cite{gal2016dropout} and mean-field variational inference (MFVI)~\cite{blundell2015weight}. Finally, parameter-space maximum a posteriori (MAP) estimation~\cite{bishop2006pattern} is also included for comparison.

\textbf{Setup.} 
We evaluate our method on four datasets: MNIST, Fashion-MNIST, CIFAR-10, and PathMNIST from MedMNIST. For OOD detection evaluation, we use standard benchmarks including FashionMNIST and NotMNIST, SVHN, as well as corrupted variants: corrupted MNIST (MNIST-C) and corrupted CIFAR-10 with severity levels 0, 2, and 4 (denoted as CIFAR-10C0/C2/C4). For MNIST and Fashion-MNIST, we use a network with two convolutional layers (32 and 64 filters of size  $ 3\times3 $ ) followed by a fully-connected layer with 128 hidden units. For CIFAR-10 and PathMNIST, we adopt a deeper architecture with six convolutional layers (32, 32, 64, 64, 128, 128 filters, all  $ 3\times3 $ ) and a final fully-connected layer of 128 units. All models are trained using the Adam optimizer, and all images are normalized to the range $ [0, 1]$. See details in~\ref{app: mnist} and~\ref{app: cifar}. For in-distribution evaluation, we report classification accuracy (ACC), negative log-likelihood (NLL), and expected calibration error (ECE). For OOD detection, we use the area under the receiver operating characteristic curve (AUROC) based on the maximum softmax probability (MSP) as the detection score. Results using alternative uncertainty-based scoring functions—namely predictive entropy and expected entropy—are provided in~\ref{app: OOD}. 

\textbf{Implementation.} For VLM-FS-EB, we use the \textit{tongyi-embedding-vision-plus} \cite{team2025tongyi} model for data embedding, producing 1,152-dimensional embedding vectors, and the \textit{qwen-vl-plus} \cite{bai2025qwen2} for context point sampling. For each training dataset, we generate 9,000 context samples. Visualisation of context examples is provided in~\ref{app: context}.  All experiments are conducted with 10 MC runs, and we report both the mean and standard deviation. Additional details on hyperparameter settings are provided in~\ref{app: Hype}.

\subsection{Performance under Standard (Full-Data) Regimes}\label{sec:E1}
This experiment evaluates the prediction performance of VLM-FS-EB, using the full training datasets. Tables~\ref{tab:main_result_1} summarises the in-distribution prediction performance across multiple datasets, including MNIST, Fashion-MNIST, CIFAR-10, and PathMNIST.  VLM-FS-EB demonstrates strong predictive performance overall: it achieves the best ACC on CIFAR-10 and the lowest NLL on CIFAR-10, and remains close to the top-performing methods across other settings. In terms of calibration (ECE), this method shows reasonable performance, though it is modestly outperformed by methods like GFSVI. 

Also, the OOD detection performance using the full training datasets is investigated in Table~\ref{tab:main_result_2}. Our VLM-FS-EB consistently achieves best or near-best performance across diverse OOD settings. In many cases, the  improvement is substantial. For example, it attains a perfect AUROC of 100.0 on MNIST vs.~NotMNIST and outperforms the strongest baseline by over 3 AUROC points. Also, the VLM-FS-EB achieves an 87.63 AUROC on the real-world medical benchmark PathMNIST vs.~BloodMNIST, surpassing all competitors by a clear margin. This demonstrates the significant potential of our method in real-world safety-critical applications.

\subsection{Performance under Limited-Data Regimes}\label{sec:E2}

To evaluate the robustness of our proposed method in data-constrained scenarios, we test all methods using only a fraction of the standard training set: specifically, 25\% of the original training data for MNIST, FashionMNIST, and CIFAR-10, and $15\%$ for PathMNIST.

The in-distribution performance is presented in Table~\ref{tab:main_result_3}. VLM-FS-EB achieves the best or second-best ACC and NLL across all datasets, with the top result in the majority of cases. Specifically, it attains the highest ACC on MNIST, FashionMNIST, and CIFAR-10, and the lowest NLL on FashionMNIST, CIFAR-10 and PathMNIST. On the remaining results (e.g., ACC on PathMNIST and NLL on MNIST), it ranks second and remains very close to the best-performing method. In terms of ECE, VLM-FS-EB generally lags behind the best-performing baselines—particularly on MNIST and CIFAR-10, where methods like MFVI and GFSVI achieve substantially lower values.

Besides, under limited-data training settings, VLM-FS-EB achieves the best OOD detection performance across all pairs as shown in Table~\ref{tab:main_result_4}. For example, it attains a near-perfect AUROC of 100.00 on MNIST vs.~NotMNIST. Also, on CIFAR-10, it consistently leads across all corruption levels (C0, 2, 4) and SVHN. Additionally, on the real-world medical benchmark PathMNIST (in-distribution) with BloodMNIST as the OOD dataset, VLM-FS-EB achieves an AUROC of 99.66, outperforming the second-best method (MFVI, 83.67) by approximately 16 AUROC points.

\subsection{Ablation Study: Role of Vision-Language Model-Derived Context Points and Embeddings}
We conduct ablation studies to assess the individual contributions of two components derived from VLMs in our framework: (1) the context points generated via our VLM-based context sampling framework; (2) the semantic embeddings used to construct the functional prior. We compare three VLM-FS-EB variants in Table~\ref{tab:main_result_5}: VLM-VLM uses VLM embeddings and VLM-generated semantic context points;
VLM-Rand uses VLM embeddings with context points randomly sampled from the training batch;
Rand-VLM uses embeddings from a randomly initialized DNN but retains VLM-generated context points.

The ablation results in Table~\ref{tab:main_result_5} demonstrate that both the VLM-derived embeddings and the VLM-generated semantic context points are crucial to the overall performance of our method. When using VLM embeddings together with VLM-generated context points (VLM-VLM), the model achieves strong in-distribution predictive performance (e.g., lowest NLL on all datasets) while simultaneously delivering the best OOD detection results across all benchmarks and scoring functions. In contrast, replacing the semantic context points with samples from the training batch (VLM-Rand) leads to a marginal improvement in accuracy on some datasets (e.g., +0.11\% on MNIST, +0.09\% on FMNIST) but causes a substantial drop in OOD detection. Conversely, for the Rand-LVM variant, in-distribution prediction  and OOD detection performance degrades significantly. These results indicate that both the VLM-derived embeddings and the VLM-generated context points are indispensable.

\section{Conclusion}\label{sec: conclusion}
This work presents a novel FS-EB regularisation framework, which introduces informative functional prior via large VLMs. Specifically, we investigate a VLM-based sampling method to produce semantically meaningful context points. Also, the informative prior of the VLM-FS-EB is formulated based on VLM embeddings. Our proposed methods are compared with various both parameter- and function-space regularisation methods. Experiment results show under the full-data regime (all training data is used), our method present competitive prediction performance and superior OOD detection results than the other baselines. By contrast, Under the limited-data regime, our method consistently outperforms the other methods in both prediction and OOD detection tasks.

\section*{Impact Statement}
This paper presents work whose goal is to advance the field of Machine
Learning. There are many potential societal consequences of our work, none
which we feel must be specifically highlighted here.

\bibliography{example_paper}
\bibliographystyle{icml2026}

%%%%%%%%%%%%%%%%%%%%%%%%%%%%%%%%%%%%%%%%%%%%%%%%%%%%%%%%%%%%%%%%%%%%%%%%%%%%%%%
%%%%%%%%%%%%%%%%%%%%%%%%%%%%%%%%%%%%%%%%%%%%%%%%%%%%%%%%%%%%%%%%%%%%%%%%%%%%%%%
% APPENDIX
%%%%%%%%%%%%%%%%%%%%%%%%%%%%%%%%%%%%%%%%%%%%%%%%%%%%%%%%%%%%%%%%%%%%%%%%%%%%%%%
%%%%%%%%%%%%%%%%%%%%%%%%%%%%%%%%%%%%%%%%%%%%%%%%%%%%%%%%%%%%%%%%%%%%%%%%%%%%%%%
\newpage
\appendix
\onecolumn
\section{Additional Details and Experiments}

\subsection{Hyperparameters}\label{app: Hype}
In Table~\ref{tab:hyperparameter_ranges}, we summarize the key hyperparameters used in VLM-FS-EB. For all tasks, models are trained with a batch size of 128 using the Adam optimizer with a learning rate of $5 \times 10^{-4}$, $\epsilon = 10^{-8}$, and momentum parameters $(\beta_1, \beta_2) = (0.9, 0.999)$. Unless otherwise specified, the number of context points is fixed to 32. Early stopping is applied in all experiments with a validation split of 0.1 and a patience of 10 epochs, with a maximum of 100 training epochs. Hyperparameter optimization is performed using randomized grid search over the defined search space. For all experiments, we conduct 60 search trials and select the optimal parameter configuration based on the minimum negative log-likelihood.
\begin{table}[htbp]
\centering
\caption{Hyperparameter Ranges}
\label{tab:hyperparameter_ranges}
\begin{tabular}{ll}
\hline
\textbf{Hyperparameter} & \textbf{Range} \\
\hline
Weight decay $\lambda$ & $\{10^{k} \mid k = -6, -5, \dots, 0\}$ \\
$\tau_1$ & $ \{10^{k} \mid k = -6, \dots, 2\}$ \\
$\tau_2$ & $ \{10^{k} \mid k = -6, \dots, 2\}$ \\
\hline
\end{tabular}
\end{table}

For all baseline methods, we employ the same optimizer, and training protocol—including batch size, learning rate schedule, early stopping criterion, and training epochs—to ensure a fair comparison. In FS-EB, the empirical prior is constructed using a randomly initialized neural network, which—following the insights of~\cite{wilson2020bayesian}. For GFSVI, we adopt a GP prior with a constant zero-mean function; the hyperparameters of the covariance kernel are learned by maximizing the log marginal likelihood on mini-batches, following the approach of Milios et al.~\cite{milios2018dirichlet}. Both FS-EB and GFSVI utilize random samples from each batch as context points to condition their functional priors. Finally, for MC Dropout, MFVI, and MAP estimation, the parameters of their isotropic Gaussian weight priors are tuned via the same randomized search procedure described above, with configurations selected based on minimal validation negative log-likelihood.

\subsection{MNIST and FashionMNIST}\label{app: mnist}
For both datasets, we employ a convolutional neural network composed of two convolutional layers with 32 and 64 filters of size $3 \times 3$, respectively. Each convolutional layer is followed by a ReLU activation and a max-pooling operation. The extracted features are then flattened and fed into a fully connected layer with 128 hidden units, followed by a final linear layer for classification. For methods using MC Dropout, dropout layers are inserted after each convolutional block and the fully connected layer, and remain active during inference to enable MCsampling. The dropout rate is set to 0.5 for both MNIST and FashionMNIST.

\subsection{CIFAR-10 and PathMNIST}\label{app: cifar}
For both datasets, we employ a convolutional neural network consisting of six convolutional layers with 32, 32, 64, 64, 128, and 128 filters of size $3 \times 3$, respectively. ReLU activations are applied after each convolutional layer. Max-pooling operations are inserted after the second, fourth, and sixth convolutional layers. The resulting feature maps are flattened and passed to a fully connected layer with 128 hidden units, followed by a final linear layer for classification.
For methods using MC Dropout, dropout layers are applied after each convolutional block and the fully connected layer, and remain active during inference to enable MC sampling. The dropout rate is set to 0.4 for CIFAR-10 and 0.2 for PathMNIST.

%\subsection{\textcolor{blue}{Visualisation of Prompt languages}}\label{app: prompt}

\subsection{Visualisation of Context Samples}\label{app: context}
Table~\ref{tab:context_visualization} illustrates representative examples of context samples generated by our VLM-based context sampling framework across four datasets. For each dataset, the in-distribution samples (left column) are real training images, while the context samples (right column) are synthetically generated by prompting a frozen vision-language model with class-specific semantic descriptions. Notably, the generated context samples maintain a consistent visual style with their corresponding in-distribution images in terms of texture, color scheme, and overall painting style. Consequently, these synthetic context samples exhibit high-level semantic coherence with the in-distribution data, thereby fulfilling our requirements for context data.

\begin{table*}[htbp]
\centering
\caption{Visualization of in-distribution samples and corresponding context images from our context point sampling pipeline.
Each image represents a $3 \times 3$ grid.}
\label{tab:context_visualization}
\begin{tabular}{c c c}
\toprule
\textbf{Dataset} & \textbf{In-distribution samples} & \textbf{Context samples} \\
\midrule
MNIST &
\includegraphics[width=0.25\linewidth]{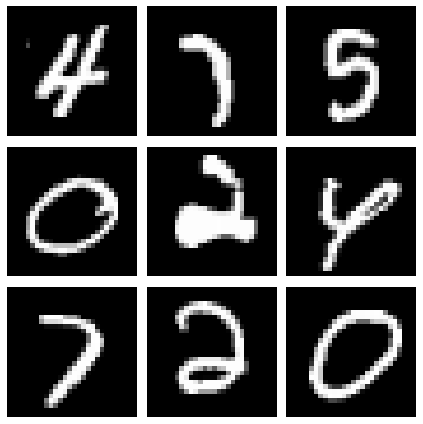} &
\includegraphics[width=0.25\linewidth]{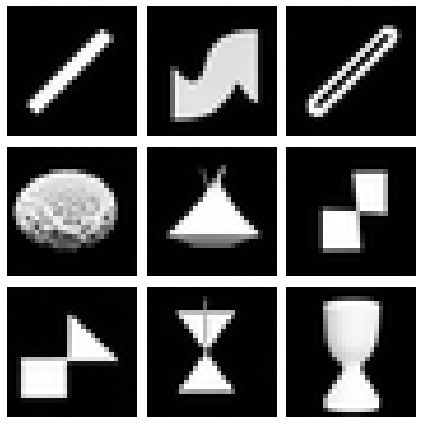} \\[6pt]
FashionMNIST &
\includegraphics[width=0.25\linewidth]{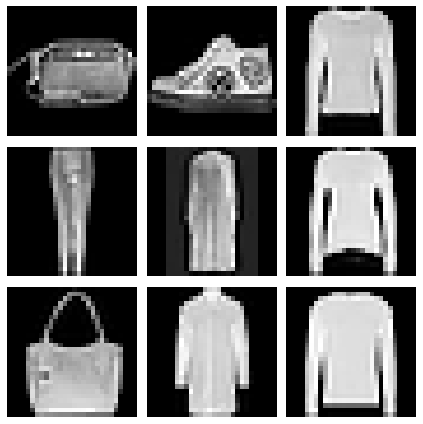} &
\includegraphics[width=0.25\linewidth]{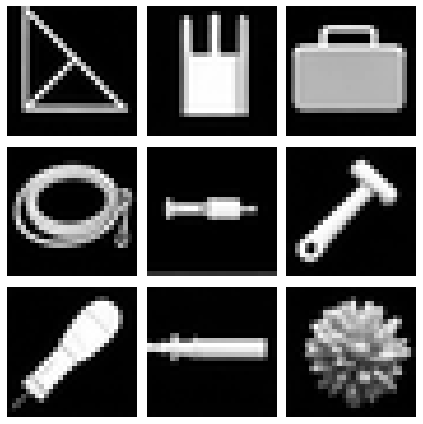} \\[6pt]

CIFAR-10 &
\includegraphics[width=0.25\linewidth]{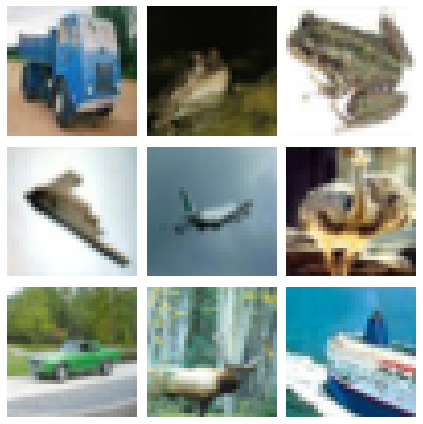} &
\includegraphics[width=0.25\linewidth]{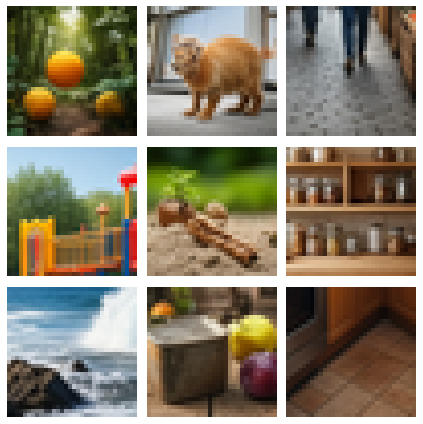} \\

PathMNIST &
\includegraphics[width=0.25\linewidth]{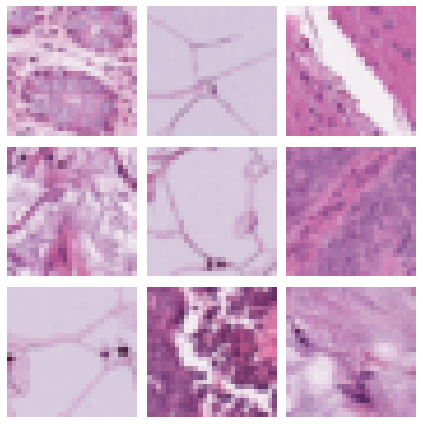} &
\includegraphics[width=0.25\linewidth]{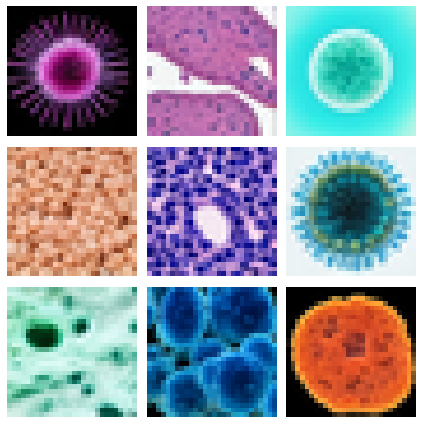} \\
\bottomrule
\end{tabular}
\end{table*}

\subsection{Resource Requirements} \label{app: time and memory}
Table~\ref{tab:Training time and memory} compares the training time and memory consumption of all methods on FMNIST and CIFAR-10. While VLM-FS-EB incurs higher computational costs than parameter-space baselines (e.g., MAP, Dropout, and MFVI), it is substantially more efficient than existing function-space regularization approaches. Specifically, VLM-FS-EB achieves a significant reduction in computational overhead: it shortens per-epoch training time by up to 48\% relative to FS-EB and by more than 66\% compared to GFSVI. Similarly, it reduces GPU memory consumption by up to 35\% against FS-EB and by over 50\% versus GFSVI. This efficiency gain primarily stems from our use of MC dropout, which imposes a sparse approximation on the posterior.

\begin{table*}[t]
\centering
\caption{Training time and memory.}
\label{tab:Training time and memory}
\resizebox{0.75\textwidth}{!}{
\begin{tabular}{l l c c c c c c}
\toprule
Metric & Dataset & VLM-FS-EB & FS-EB & GFSVI & Dropout & MFVI & MAP \\
\midrule

\multirow{2}{*}{Time(s/epoch) $\downarrow$}
& FMNIST  & 6.65 & 12.86 & 20.21 & 4.74 & 8.04 & 7.04 \\
& CIFAR-10 & 10.72 & 19.30 & 26.06 & 8.45 & 13.69 & 7.70 \\
\midrule

\multirow{2}{*}{Memory (MB) $\downarrow$}
& FMNIST  & 524.19 & 810.75 & 1059.31 & 437.53 & 634.57 & 65.27 \\
& CIFAR-10 & 1657.11 & 2154.40 & 2604.89 & 1274.19 & 1714.41 & 112.44 \\
\bottomrule
\end{tabular}
}
\end{table*}

\subsection{OOD Detection with Alternative Uncertainty Scoring Functions}\label{app: OOD}

We present comprehensive OOD detection results across four benchmark datasets—MNIST, FashionMNIST, CIFAR-10, and PathMNIST—under two distinct data regimes: (i) full training data (Tables~\ref{tab:OOD_1} and~\ref{tab:OOD_2}) and (ii) partial training data (Tables~\ref{tab:OOD_3} and~\ref{tab:OOD_4}). In all cases, we report AUROC scores using two standard uncertainty scoring functions: the entropy of the softmax output and the expected predictive entropy under the approximate posterior. The experimental protocols for the full-data and partial-data settings strictly follow Sections~\ref{sec:E1} and~\ref{sec:E2}, respectively.

Under the full-data regime (Tables~\ref{tab:OOD_1} and~\ref{tab:OOD_2}), VLM-FS-EB consistently achieves best OOD detection performance across most dataset pairs and both scoring functions. Notably, it attains perfect AUROC (100.00) on MNIST vs. FMNIST/NotMNIST. These findings align closely with the results reported in Section~\ref{sec:E1}, where our method also demonstrates superior OOD performance.

Also, under the partial-data regime (Tables~\ref{tab:OOD_3}~\ref{tab:OOD_4}), VLM-FS-EB consistently outperforms all baseline methods across most evaluated benchmarks. Especially, on PathMNIST, it achieves an AUROC of 99.42, compared to 86.32 for the best-performing baseline. These results indicate that incorporating semantic knowledge from frozen vision-language models can effectively mitigate the challenges posed by limited training data, leading to improved uncertainty quantification in data-scarce settings.

\begin{table*}[t]
\centering
\caption{OOD detection performance under full training data. We report AUROC scores computed using the entropy of the softmax output as the OOD detection score. \textbf{Best} results are highlighted in bold and \underline{second best} results are underlined.
}
\label{tab:OOD_1}
\resizebox{\textwidth}{!}{
\begin{tabular}{l l c c c c c c}
\toprule
In & OOD Dataset & VLM-FS-EB &  FS-EB & GFSVI & Dropout & MFVI & MAP \\
\midrule

\multirow{4}{*}{MNIST}
& FMNIST & \textbf{100.00  $  \pm  $  0.003} & \underline{99.33  $  \pm  $  0.131} & 98.76  $  \pm  $  0.606 & 99.02  $  \pm  $  0.302 & 99.00  $  \pm  $  0.191 & 98.93  $  \pm  $  0.240  \\
& NotMNIST & \textbf{100.00  $  \pm  $  0.000} & 95.39  $  \pm  $  1.037 & 91.33  $  \pm  $  3.925 & \underline{97.10  $  \pm  $  0.300} & 94.15  $  \pm  $  0.802 & 95.65  $  \pm  $  0.657 \\
& MNIST-c & \textbf{92.44  $  \pm  $  0.962} & \underline{85.69  $  \pm  $  1.109} & 84.13  $  \pm  $  3.377 & 85.00  $  \pm  $  0.674 & 82.99  $  \pm  $  0.950 & 82.51  $  \pm  $  0.744 \\

\midrule

\multirow{2}{*}{FMNIST}
& MNIST & \underline{88.89  $  \pm  $  1.884} & 86.85  $  \pm  $  2.918 & \textbf{90.35  $  \pm  $  1.864} & 83.69  $  \pm  $  1.761 & 79.40  $  \pm  $  3.055 & 79.61  $  \pm  $  2.405 \\
& NotMNIST & \textbf{89.99  $  \pm  $  1.513} & 79.12  $  \pm  $  3.969 & 75.67  $  \pm  $  2.022 & \underline{82.53  $  \pm  $  1.654} & 71.77  $  \pm  $  4.782 & 69.91  $  \pm  $  1.589 \\

\midrule

\multirow{4}{*}{CIFAR-10}
& SVHN & \textbf{86.32  $  \pm  $  1.550} & 81.13  $  \pm  $  2.575 & 78.73  $  \pm  $  1.045 & \underline{81.87  $  \pm  $  3.500} & 81.77  $  \pm  $  1.892 & 80.81  $  \pm  $  1.782 \\
& CIFAR-10C0 & \textbf{75.66  $  \pm  $  1.776} & 68.61  $  \pm  $  1.957 & 61.83  $  \pm  $  1.075 & \underline{69.10  $  \pm  $  1.360} & 68.61  $  \pm  $  1.477 & 64.75  $  \pm  $  1.681 \\
& CIFAR-10C2 & \textbf{65.24  $  \pm  $  1.302} & \underline{63.50  $  \pm  $  1.065} & 57.09  $  \pm  $  0.721 & 62.72  $  \pm  $  0.843 & 63.26  $  \pm  $  0.901 & 59.72  $  \pm  $  1.042 \\
& CIFAR-10C4 & 54.58  $  \pm  $  0.573 & \textbf{55.73  $  \pm  $  0.485} & 52.74  $  \pm  $  0.372 & 54.88  $  \pm  $  0.423 & \underline{55.54  $  \pm  $  0.455} & 53.67  $  \pm  $  0.439 \\

\midrule
\multirow{1}{*}{PathMNIST}
& BloodMNIST & \textbf{90.94  $  \pm  $  10.239} & 84.09  $  \pm  $  11.349 & 74.93  $  \pm  $  6.942 & 71.63  $  \pm  $  15.708 & \underline{86.41  $  \pm  $  8.434} & 73.30  $  \pm  $  13.452 \\

\bottomrule
\end{tabular}
}
\end{table*}

\begin{table*}[t]
\centering
\caption{OOD detection performance under full training data. We report AUROC scores computed using the expected entropy as the OOD detection score. \textbf{Best} results are highlighted in bold and \underline{second best} results are underlined.
}
\label{tab:OOD_2}
\resizebox{\textwidth}{!}{
\begin{tabular}{l l c c c c c c}
\toprule
In & OOD Dataset & VLM-FS-EB &  FS-EB & GFSVI & Dropout & MFVI & MAP \\
\midrule
\multirow{3}{*}{MNIST}
& FMNIST & \textbf{100.00  $  \pm  $  0.002} & \underline{99.35  $  \pm  $  0.124} & 98.78  $  \pm  $  0.605 & 99.20  $  \pm  $  0.253 & 99.06  $  \pm  $  0.184 & 98.93  $  \pm  $  0.240 \\
& NotMNIST & \textbf{100.00  $  \pm  $  0.002} & 94.92  $  \pm  $  1.044 & 91.23  $  \pm  $  3.955 & \underline{96.72  $  \pm  $  0.338} & 93.39  $  \pm  $  0.728 & 95.65  $  \pm  $  0.657 \\
& MNIST-c & \textbf{92.78  $  \pm  $  0.961} & \underline{85.80  $  \pm  $  1.127} & 84.14  $  \pm  $  3.380 & 85.43  $  \pm  $  0.666 & 83.19  $  \pm  $  0.948 & 82.51  $  \pm  $  0.744 \\

\midrule

\multirow{2}{*}{FMNIST}
& MNIST & \underline{85.03  $  \pm  $  2.192} & 84.02  $  \pm  $  2.935 & \textbf{90.15  $  \pm  $  1.869} & 74.22  $  \pm  $  1.961 & 75.57  $  \pm  $  2.957 & 79.61  $  \pm  $  2.405 \\
& NotMNIST & \textbf{84.84  $  \pm  $  1.922} & \underline{75.77  $  \pm  $  3.735} & 75.32  $  \pm  $  2.004 & 73.40  $  \pm  $  1.810 & 67.89  $  \pm  $  3.990 & 69.91  $  \pm  $  1.589 \\

\midrule

\multirow{4}{*}{CIFAR-10}
& SVHN & \textbf{88.66  $  \pm  $  1.500} & 83.72  $  \pm  $  2.713 & 80.15  $  \pm  $  1.077 & \underline{85.57  $  \pm  $  3.525} & 84.66  $  \pm  $  2.110 & 80.81  $  \pm  $  1.782 \\
& CIFAR-10C0 & \textbf{76.54  $  \pm  $  1.851} & 68.75  $  \pm  $  2.180 & 61.99  $  \pm  $  1.067 & \underline{69.55  $  \pm  $  1.490} & 68.62  $  \pm  $  1.636 & 64.75  $  \pm  $  1.681 \\
& CIFAR-10C2 & \textbf{65.71  $  \pm  $  1.381} & \underline{63.74  $  \pm  $  1.205} & 57.24  $  \pm  $  0.711 & 63.04  $  \pm  $  0.899 & 63.53  $  \pm  $  1.014 & 59.72  $  \pm  $  1.042 \\
& CIFAR-10C4 & 54.59  $  \pm  $  0.591 & \textbf{55.93  $  \pm  $  0.538} & 52.84  $  \pm  $  0.377 & 55.03  $  \pm  $  0.426 & \underline{55.81  $  \pm  $  0.481} & 53.67  $  \pm  $  0.439 \\

\midrule

\multirow{1}{*}{PathMNIST}
& BloodMNIST & \textbf{92.42  $  \pm  $  9.054} & 84.09  $  \pm  $  11.349 & 74.93  $  \pm  $  6.942 & 71.63  $  \pm  $  15.708 & \underline{86.41  $  \pm  $  8.434} & 73.30  $  \pm  $  13.452 \\

\bottomrule
\end{tabular}
}
\end{table*}

\begin{table*}[t]
\centering
\caption{OOD detection performance under partial training data settings. We report AUROC scores computed using entropy of the softmax output as the OOD detection score. \textbf{Best} results are highlighted in bold and \underline{second best} results are underlined.
}
\label{tab:OOD_3}
\resizebox{\textwidth}{!}{
\begin{tabular}{l l c c c c c c}
\toprule
In & OOD Dataset & VLM-FS-EB &  FS-EB & GFSVI & Dropout & MFVI & MAP \\
\midrule

\multirow{3}{*}{MNIST}
& FMNIST & \textbf{100.00  $  \pm  $  0.002} & \underline{99.23  $  \pm  $  0.136} & 97.71  $  \pm  $  0.747 & 98.95  $  \pm  $  0.253 & 97.10  $  \pm  $  1.305 & 98.02  $  \pm  $  0.512 \\
& NotMNIST & \textbf{100.00  $  \pm  $  0.001} & 95.76  $  \pm  $  0.682 & 88.98  $  \pm  $  5.134 & \underline{96.62  $  \pm  $  0.265} & 90.45  $  \pm  $  3.142 & 93.44  $  \pm  $  1.377 \\
& MNIST-c & \textbf{91.94  $  \pm  $  0.916} & \underline{84.68  $  \pm  $  0.983} & 75.82  $  \pm  $  2.567 & 82.61  $  \pm  $  0.617 & 79.62  $  \pm  $  1.686 & 80.56  $  \pm  $  0.914 \\

\midrule
\multirow{2}{*}{FMNIST}
& MNIST & \textbf{89.43  $  \pm  $  1.944} & 86.47  $  \pm  $  3.921 & \underline{87.47  $  \pm  $  2.530} & 85.94  $  \pm  $  2.204 & 80.26  $  \pm  $  5.863 & 72.61  $  \pm  $  2.044 \\
& NotMNIST & \textbf{88.86  $  \pm  $  1.156} & 79.17  $  \pm  $  3.406 & 77.67  $  \pm  $  2.172 & \underline{82.96  $  \pm  $  2.660} & 74.06  $  \pm  $  3.612 & 66.25  $  \pm  $  2.473 \\

\midrule

\multirow{4}{*}{CIFAR-10}
& SVHN & \textbf{78.93  $  \pm  $  2.576} & 73.55  $  \pm  $  0.836 & 71.59  $  \pm  $  4.307 & 72.77  $  \pm  $  3.290 & 71.35  $  \pm  $  3.230 & \underline{76.83  $  \pm  $  3.475} \\
& CIFAR-10C0 & \textbf{73.12  $  \pm  $  2.001} & 57.10  $  \pm  $  1.213 & 56.65  $  \pm  $  2.721 & \underline{62.74  $  \pm  $  2.908} & 56.53  $  \pm  $  1.064 & 58.49  $  \pm  $  1.972 \\
& CIFAR-10C2 & \textbf{62.91  $  \pm  $  1.691} & 54.56  $  \pm  $  0.823 & 54.01  $  \pm  $  1.751 & \underline{58.08  $  \pm  $  2.054} & 54.37  $  \pm  $  0.671 & 55.35  $  \pm  $  1.225 \\
& CIFAR-10C4 & \textbf{53.23  $  \pm  $  0.696} & 52.14  $  \pm  $  0.474 & 51.63  $  \pm  $  0.965 & \underline{53.08  $  \pm  $  0.855} & 52.08  $  \pm  $  0.384 & 52.31  $  \pm  $  0.559 \\

\midrule
\multirow{1}{*}{PathMNIST}
& BloodMNIST & \textbf{99.42  $  \pm  $  0.042} & \underline{86.32  $  \pm  $  11.365} & 58.26  $  \pm  $  6.735 & 42.72  $  \pm  $  25.745 & 86.16  $  \pm  $  8.737 & 58.56  $  \pm  $  17.860 \\

\bottomrule
\end{tabular}
}
\end{table*}

\begin{table*}[t]
\centering
\caption{OOD detection performance under partial training data settings. We report AUROC scores computed using expected entropy as the OOD detection score. Results are reported as mean $\pm$ standard deviation over 10 MCruns.
\textbf{Best} results are highlighted in bold and \underline{second best} results are underlined.
}
\label{tab:OOD_4}
\resizebox{\textwidth}{!}{
\begin{tabular}{l l c c c c c c}
\toprule
In & OOD Dataset & VLM-FS-EB &  FS-EB & GFSVI & Dropout & MFVI & MAP \\

\midrule

\multirow{3}{*}{MNIST}
& FMNIST & \textbf{100.00  $  \pm  $  0.001} & \underline{99.33  $  \pm  $  0.139} & 97.70  $  \pm  $  0.710 & 99.27  $  \pm  $  0.185 & 97.24  $  \pm  $  1.332 & 98.02  $  \pm  $  0.512 \\
& NotMNIST & \textbf{100.00  $  \pm  $  0.001} & 95.37  $  \pm  $  0.674 & 88.10  $  \pm  $  5.265 & \underline{96.50  $  \pm  $  0.347} & 89.92  $  \pm  $  3.172 & 93.44  $  \pm  $  1.377 \\
& MNIST-c & \textbf{92.36  $  \pm  $  0.907} & \underline{85.03  $  \pm  $  1.014} & 75.93  $  \pm  $  2.503 & 83.35  $  \pm  $  0.604 & 79.79  $  \pm  $  1.711 & 80.56  $  \pm  $  0.914 \\

\midrule

\multirow{2}{*}{FMNIST}
& MNIST & 82.76  $  \pm  $  2.424 & \underline{83.41  $  \pm  $  3.014} & \textbf{87.23  $  \pm  $  2.544} & 78.00  $  \pm  $  2.007 & 75.02  $  \pm  $  4.203 & 72.61  $  \pm  $  2.044 \\
& NotMNIST & \textbf{81.82  $  \pm  $  1.379} & 75.76  $  \pm  $  3.372 & \underline{76.53  $  \pm  $  2.225} & 74.11  $  \pm  $  2.797 & 68.91  $  \pm  $  3.738 & 66.25  $  \pm  $  2.473 \\

\midrule

\multirow{4}{*}{CIFAR-10}
& SVHN & \textbf{81.54  $  \pm  $  2.551} & 75.75  $  \pm  $  0.833 & 73.15  $  \pm  $  4.632 & \underline{77.28  $  \pm  $  3.068} & 73.40  $  \pm  $  3.949 & 76.83  $  \pm  $  3.475 \\
& CIFAR-10C0 & \textbf{74.10  $  \pm  $  2.024} & 57.57  $  \pm  $  1.383 & 56.87  $  \pm  $  2.864 & \underline{63.79  $  \pm  $  3.021} & 56.94  $  \pm  $  1.272 & 58.49  $  \pm  $  1.972 \\
& CIFAR-10C2 & \textbf{63.42  $  \pm  $  1.803} & 54.89  $  \pm  $  0.966 & 54.18  $  \pm  $  1.809 & \underline{58.95  $  \pm  $  2.190} & 54.67  $  \pm  $  0.814 & 55.35  $  \pm  $  1.225 \\
& CIFAR-10C4 & \underline{53.18  $  \pm  $  0.776} & 52.27  $  \pm  $  0.551 & 51.70  $  \pm  $  0.970 & \textbf{53.49  $  \pm  $  0.949} & 52.22  $  \pm  $  0.409 & 52.31  $  \pm  $  0.559 \\

\midrule
\multirow{1}{*}{PathMNIST}
& BloodMNIST & \textbf{99.42  $  \pm  $  0.056} & 77.87  $  \pm  $  13.171 & 54.95  $  \pm  $  6.386 & 40.97  $  \pm  $  25.016 & \underline{78.86  $  \pm  $  8.943} & 58.56  $  \pm  $  17.860 \\

\bottomrule
\end{tabular}
}
\end{table*}

\end{document}